\documentclass{article}
\usepackage[preprint]{spconf,amsmath,graphicx,amssymb,circledsteps}
\usepackage{multirow}
\usepackage{booktabs}
\usepackage{cite}

\copyrightnotice{\copyright\ IEEE 2021}
\toappear{To appear in {\it Proc.\ ASRU2021, December 13-17, 2021, Cartagena, Colombia}}
\title{Tree-constrained Pointer Generator for End-to-end\\ Contextual Speech Recognition}
\name{Guangzhi Sun, Chao Zhang, Philip C. Woodland\thanks{Guangzhi Sun is funded by Cambridge Trust.}}
\address{Cambridge University Engineering Dept., Trumpington St., Cambridge, CB2 1PZ U.K.\\
\small{\texttt{\{gs534,cz277,pcw\}@eng.cam.ac.uk}}}
\begin{document}
\ninept
\maketitle
\begin{abstract}
Contextual knowledge is important for real-world automatic speech recognition (ASR) applications. 
In this paper, a novel tree-constrained pointer generator (TCPGen) component is proposed that incorporates such knowledge as a list of biasing words into both attention-based encoder-decoder and transducer end-to-end ASR models in a neural-symbolic way. TCPGen structures the biasing words into an efficient prefix tree to serve as its symbolic input and creates a neural shortcut between the tree and the final ASR output distribution to facilitate recognising biasing words during decoding. Systems were trained and evaluated on the Librispeech corpus where biasing words were extracted at the scales of an utterance, a chapter, or a book to simulate different application scenarios.  
Experimental results showed that TCPGen consistently improved word error rates (WERs) compared to the baselines, and in particular, achieved significant WER reductions on the biasing words. TCPGen is highly efficient: it can handle 5,000 biasing words and distractors and only add a small overhead to memory use and computation cost. 




\end{abstract}
%
\section{Introduction}
\label{sec:intro}
Contextual biasing, which integrates contextual knowledge into an automatic speech recognition (ASR) system, has become increasingly important to many applications \cite{shallow_context_1,shallow_context_2,shallow_context_3,deep_context_1,deep_context_2,deep_context_3,deep_context_4,deep_context_5,deepshallow,DBRNNT,ne_correction,unsupervised_context,dialogue_contextual,word_mapping,lecture_context,dialogue_contextual_2,audiovisual_context,lm_pointer}. Contextual knowledge is often represented by a list (referred to as a \textit{biasing list}) of words or phrases (referred to as \textit{biasing words}) that are likely to appear in an utterance in a given context. Examples of resources to find biasing lists include a user's contact book or playlist, recently visited websites, and the ontology of a dialogue system \textit{etc}. Despite the fact that such biasing words may only have a small impact on the overall word error rate (WER), these words are mostly rare content words, frequently nouns or proper nouns, that are important to downstream tasks and are thus highly valuable.
Moreover, a word is more likely to be accurately recognised when it is incorporated in the biasing list, which therefore improves the controllability of an ASR system.
However, since contextual knowledge is specific to the dynamic test-time context, it is not easy to be fully integrated during model training. This is particularly difficult for end-to-end systems \cite{e2e_attention_1, e2e_rnnt_1}, which are often designed to use a single static model to integrate all of the required knowledge. Therefore, dedicated biasing methods have been developed, including shallow fusion (SF) with a special weighted finite-state transducer (WFST) or language model (LM) adapted for the contextual knowledge \cite{shallow_context_1,shallow_context_2,shallow_context_3,lm_pointer,unsupervised_context,word_mapping} and attention-based deep context approaches \cite{deep_context_1,deep_context_2,deep_context_3,deep_context_4,deep_context_5}. Recently, deep biasing (DB) was proposed to deal with large biasing lists efficiently by organising the biasing words into a prefix tree and feeding it to model output layers \cite{deepshallow,DBRNNT}.

In this paper, a tree-constrained pointer generator (TCPGen) component is proposed to improve end-to-end ASR models by enabling them to leverage external contextual knowledge.
At each inference step, TCPGen not only estimates an extra output probability distribution (denoted as the TCPGen distribution) based on the contextual knowledge and the decoder state, but also predicts a generation probability to form the final output distribution by interpolating the TCPGen distribution with the original output distribution. 
In this way, TCPGen creates a neural shortcut between the biasing lists and the ASR output distributions, which facilitates the spotting of involved biasing words using a single neural network model trained in an end-to-end fashion. To the best of authors' knowledge, this is the first work that introduces the idea of pointer generators \cite{pointer_1} into end-to-end ASR to help address the issue of external knowledge integration. 
In addition, TCPGen resolves two major shortcomings of the original pointer generator approach \cite{pointer_1,pointer_2,pointer_3}. 
First, instead of fully relying on an attention mechanism to directly attend to all biasing words which tends to be memory intensive 
for large biasing lists, TCPGen organises the biasing words into a symbolic prefix tree and only attends to a subset of the biasing words at each time step in decoding. 
Second, by building prefix trees with subword units, TCPGen consequently enables the final ASR output distributions to be subword-unit-based, while with the original pointer generators \cite{pointer_1,pointer_2,pointer_3,lm_pointer}, the output distributions have to be whole-word-based. 
As a result, TCPGen combines the advantages of both neural and symbolic methods, and
improves pointer generators so that they can be applied in ASR applications with large biasing lists and subword output units.

As a generic method, TCPGen is integrated into both attention-based encoder-decoder (AED) \cite{e2e_attention_1,e2e_attention_2,e2e_attention_3,e2e_attention_4,e2e_attention_5,LAS} and recurrent neural network transducer (RNN-T) \cite{e2e_rnnt_1,e2e_rnnt_2,e2e_rnnt_3,e2e_rnnt_4,e2e_rnnt_5} end-to-end ASR models in this paper. 
Experiments were performed on Librispeech audiobook data. To simulate the use of contextual knowledge in practice, biasing lists were organised for each utterance following \cite{DBRNNT} which was shown 
to be a valid simulation. Additionally, 
chapter-level and book-level biasing lists were also arranged to simulate a more realistic scenario where biasing lists can be inferred from a large amount of text knowledge such as presentations and conferences. Improvements in word error rate (WER) were achieved by using TCPGen in both AED and RNN-T on test sets compared to the baseline DB and SF methods, with a significant WER reduction on the biasing words. 


The rest of the paper is organised as follows: 
Sec.~\ref{sec:relwork} reviews 
related contextual biasing and end-to-end ASR work. 
Sec.~\ref{sec:TCPGen} gives details of the TCPGen technique and its integration with AED and RNN-T. Sec.~\ref{sec:setup} describes our experimental setup, followed by a discussion of the results in Sec.~\ref{sec:result}. 
Finally, conclusions are 
in Sec.~\ref{sec:conclusion}.

\vspace{-0.2cm}
\section{Background}
\label{sec:relwork}
\vspace{-0.2cm}
\subsection{End-to-end ASR Models}
\label{sec:e2easr}
End-to-end ASR approaches including AED \cite{e2e_attention_1,e2e_attention_2,e2e_attention_3,e2e_attention_4,e2e_attention_5,e2e_attention_6,e2e_attention_7,e2e_attention_8,LAS}, RNN-T \cite{e2e_rnnt_1,e2e_rnnt_2,e2e_rnnt_3,e2e_rnnt_4,e2e_rnnt_5} and others \cite{ISCA,e2e_ctc_1,e2e_ctc_2,joint_ctc_atten,e2eLFMMI,SJTU}, have become increasingly popular due to their simplicity and competitive performance. This section introduces various terms and the notation used for AED and RNN-T. Both systems directly optimise the posterior distribution $P(y_{1:N}|\mathbf{x}_{1:T})$ over the sequence of output subword units, $y_{1:N}$, given the input acoustic feature sequence $\mathbf{x}_{1:T}$.

\vspace{-0.3cm}
\subsubsection{Attention-based Encoder-Decoder}
A standard AED contains three components: an \textit{encoder}, a \textit{decoder} and an \textit{attention network}. 
The encoder 
encodes the input, $\mathbf{x}_{1:T}$, into a sequence of high-level features, $\mathbf{h}^{\text{enc}}_{1:T}$. 
At each decoding step $i$, 
an attention mechanism is first used to combine the encoder output sequence into a single context vector, $\mathbf{c}_i$ as part of the input to the decoder. 
The decoder computation is thus as follows:
\vspace{-0.1cm}
\begin{equation}
    \mathbf{h}^{\text{dec}}_i = \text{Decoder}(\mathbf{y}_{i-1}, \mathbf{h}^{\text{dec}}_{i-1}, \mathbf{c}_{i}),
    \vspace{-0.1cm}
\end{equation}
where Decoder$(\cdot)$ denotes the decoder network and $\mathbf{y}_{i-1}$ is the embedding of the preceding subword unit. The posterior distribution can be estimated with a Softmax output layer.
\vspace{-0.1cm}
\begin{equation}
    P(y_i | y_{1:i-1}, \mathbf{x}_{1:T}) = \text{Softmax}(\mathbf{W}^{\text{O}}[\mathbf{h}^{\text{dec}}_i; \mathbf{c}_{i}]),
    \label{eq:lasoutput}
    \vspace{-0.1cm}
\end{equation}
where $[\cdot;\cdot]$ denotes the concatenation of two vectors. 
At inference time, the recognition result $y^*_{1:N}$ is approximately calculated by performing \textit{beam search}.
Moreover, LM shallow fusion \cite{lmfusion_2,lmfusion_3,lmfusion_5,ilme_1,ilmt_1,HAT} can be performed via a log-linear combination as shown in Eqn. \eqref{eq:shallowfusion}.
\vspace{-0.1cm}
\begin{equation}
    {y}^*_{1:N} = \underset{{y}_{1:N}}{\text{arg max}}\,\log P({y}_{1:N}|\mathbf{x}_{1:T}) + \lambda\,\log P^{\text{LM}} (y_{1:N}),
    \label{eq:shallowfusion}
    \vspace{-0.2cm}
\end{equation}
where $\lambda$ is a hyper-parameter that controls the relative importance of the LM output probabilities $P^{\text{LM}} (y_{1:N})$.

\vspace{-0.2cm}
\subsubsection{RNN-Transducer}
RNN-T is widely used in streaming ASR tasks. A standard RNN-T consists of an encoder, a predictor and a joint network. The encoder performs a similar task to the one in AED whose output is $\mathbf{h}^{\text{enc}}_{1:T}$. The predictor encodes all subword units in the history into a vector, $\mathbf{h}^{\text{pred}}_i$, analogous to an LM.
Given $K$ encoder outputs and $N$ predictor outputs, the joint network determines the output distribution $P(z_{i,t} | y_{1:i}, \mathbf{h}^{\text{enc}}_{t})$ for each combination of $i$ and $t$ as shown in Eqns. \eqref{eq:joint} and \eqref{eq:joint2}.
\begin{equation}
    \mathbf{h}^{\text{joint}}_{i, t} = \tanh(\mathbf{W}^{\text{joint}} [\mathbf{h}^{\text{pred}}_i; \mathbf{h}^{\text{enc}}_{t}]),
    \label{eq:joint}
\end{equation}
\begin{equation}
    P(z_{i,t} | y_{1:i-1}, \mathbf{h}^{\text{enc}}_{t}) = \text{Softmax}(\mathbf{W}^{\text{joint}}_2\mathbf{h}^{\text{joint}}_{i, t}),
    \label{eq:joint2}
\end{equation}
where $\mathbf{W}^{\text{joint}}$'s are parameter matrices of the joint network. The space of output $z$ is the union of all subword units and a blank symbol $\varnothing$ denoting no output subword unit. 
The complete set of $T \times N$ decisions jointly determines the distribution over all alignments and the posterior distribution is calculated by marginalising over these alignments.
During inference, the best alignment from beam search will be used as the recognition result. LM shallow fusion in Eqn. (\ref{eq:shallowfusion}) can also be applied to RNN-T only when the previous output is not $\varnothing$, as the LM does not have a probability for $\varnothing$.

\subsection{Contextual Biasing}
Various contextual biasing approaches have been developed over the past few years. One common approach is to present the biasing list as an extra weighted finite-state transducer and incorporate it into a class-based LM with shallow fusion \cite{shallow_context_1,shallow_context_2,shallow_context_3,word_mapping}. However, this usually relies on special context prefixes like ``call'' or ``play'', and may not be flexible enough. Meanwhile, the attention-based deep context approach uses an attention component to encode the list of biasing words into a vector as part of the input to an AED or RNN-T \cite{deep_context_1,deep_context_2,deep_context_3,deep_context_4,deep_context_5}. Despite the direct injection of contextual knowledge into models without using special prefixes, this method becomes more memory intensive and less effective when more biasing words are involved. 

Work in \cite{deepshallow} proposed DB which combined the deep context and WFST methods together in an RNN-T. It also improves the efficiency by extracting the biasing vector from only a subset of word pieces constrained by a prefix tree representing the biasing list. Work in \cite{DBRNNT} extended the prefix-tree-based method to RNN LMs which are used for shallow fusion to achieve further improvements in biasing words. Moreover, while previous studies only focused on industry data sets, researchers in \cite{DBRNNT} proposed and justified a simulation of the contextual biasing task on open source data by adding a large number of distractors to the list of biasing words in an utterance.

\section{Tree-constrained Pointer Generator}
\label{sec:TCPGen}
\begin{figure}[t]
    \centering
    \includegraphics[scale=0.37]{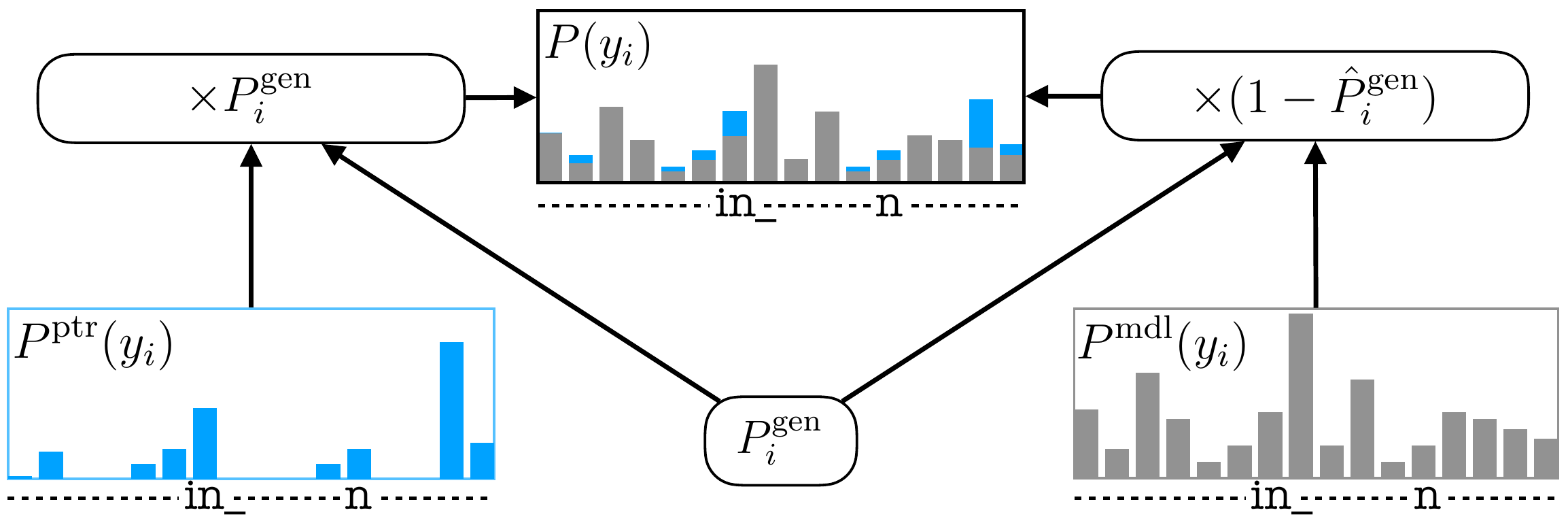}
    \caption{Illustration of interpolation in TCPGen with corresponding terms in Eqn. (\ref{eq:TCPGen_final}). $P^\text{ptr}(y_i)$ is the TCPGen distribution. $P^\text{mdl}(y_i)$ is the distribution from a standard end-to-end model. $P(y_i)$ is the final output distribution. $\hat{P}^\text{gen}_i$ and $P^\text{gen}_i$ are the scaled and unscaled generation probabilities.}
    \label{fig:TCPGen_illustrate}
    \vspace{-0.3cm}
\end{figure}

TCPGen is a neural network-based component that can be integrated into a standard 
AED or RNN-T to form a single model for end-to-end joint training. 
TCPGen calculates a distribution over all valid subword units constrained by a prefix tree built based on the biasing list. 
TCPGen also predicts a generation probability indicating how much contextual biasing is needed to decode the current token. 
The final output distribution is the weighted sum of the TCPGen distribution and the original AED or RNN-T output distribution
(Fig.~\ref{fig:TCPGen_illustrate}). 

\begin{figure}[t]
    \centering
    \includegraphics[scale=0.36]{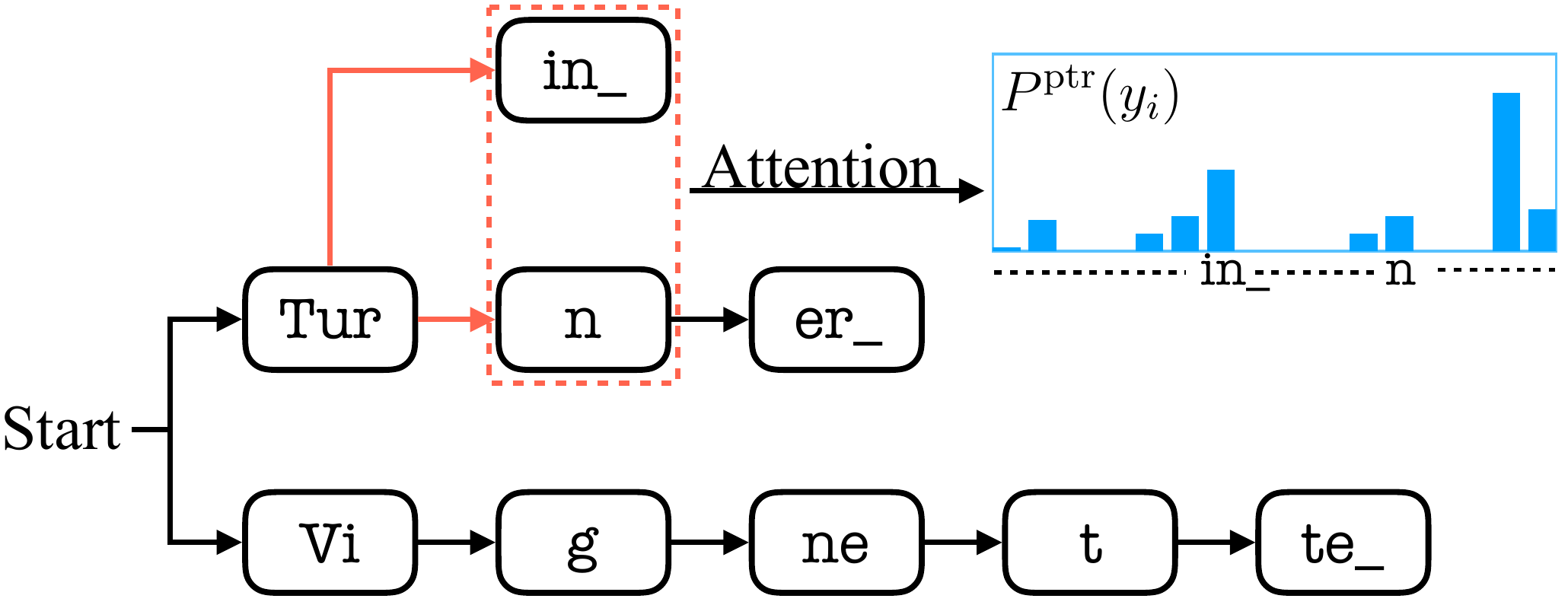}
    \vspace{-0.2cm}
    \caption{An example of prefix tree search and attention in TCPGen. With previous output \texttt{Tur}, \texttt{in\_} and \texttt{n} are two valid word pieces on which attention will be performed. A word end unit is denoted by \texttt{\_}.}
    \label{fig:trie}
\end{figure}

The prefix tree is a key symbolic representation of
the external contextual knowledge in TCPGen. For simplicity, examples and equations in this section are presented for a specific search path, which can be generalised easily to beam-search with multiple paths. For the example of prefix tree shown in Fig. \ref{fig:trie}., when decoding with the previous token $y_{i-1}=$\texttt{Tur}, based on the biasing list, a subset of word pieces $\mathcal{Y}^{\text{tree}}_i$ are valid, which includes \texttt{n} and \texttt{in\_} shown in the example. 
Denoting $\mathbf{q}_i$ as the query vector carrying the current decoding history and acoustic information, $\mathbf{k}_j$ as the embedding  of the $j$-th word piece, $\mathbf{K}=[...,\mathbf{k}_j,...]$ as the key vectors, a scaled dot-product attention is performed between $\mathbf{q}_i$ and $\mathbf{K}$ to compute a TCPGen output vector $\mathbf{h}^{\text{ptr}}_i$ as shown in Eqn. \eqref{eq:TCPGen_attention} and Eqn. \eqref{eq:TCPGen_value}.
\vspace{-0.1cm}
\begin{equation}
    P^{\text{ptr}}(y_{i}|y_{1:i-1},\mathbf{x}_{1:T}) = \text{Softmax}(\text{Mask}(\mathbf{q}_i\mathbf{K}^\text{T}/\sqrt{d})),
    \label{eq:TCPGen_attention}
    \vspace{-0.1cm}
\end{equation}
\begin{equation}
    \mathbf{h}^{\text{ptr}}_i = \sum\nolimits_{j} P^{\text{ptr}}(y_i=j|y_{1:i-1},\mathbf{x}_{1:T})\,\mathbf{v}^\text{T}_j,
    \label{eq:TCPGen_value}
    \vspace{-0.1cm}
\end{equation}
where $d$ is the size of $\mathbf{q}_i$ (see \cite{transformer}), Mask$(\cdot)$ is to mask out word pieces that are not in $\mathcal{Y}^{\text{tree}}_i$, and $\mathbf{v}_j$ is the value vector relevant 
to $j$.

For more flexibility, an \textit{out-of-list} (OOL) token is included in $\mathcal{Y}^{\text{tree}}_i$ indicating that the query is not able to find any valid word piece in the biasing list. 
To ensure that the sum of the final distribution is 1, the generator probability is scaled as $\hat{P}^\text{gen}_i=P^\text{gen}_i(1-P^\text{ptr}(\text{OOL}))$ to exclude the OOL probability from the generation probability before interpolation. Then, the interpolation can be written as Eqn. \eqref{eq:TCPGen_final}.
\begin{equation}
    P(y_i) = P^{\text{mdl}}(y_i)(1-\hat{P}^\text{gen}_i) + P^{\text{ptr}}(y_i)P^\text{gen}_i,
    \label{eq:TCPGen_final}
\end{equation}
where conditions, $y_{1:i-1}, \mathbf{x}_{1:T}$, are omitted for clarity. $P^{\text{mdl}}(y_i)$ represents the output distribution from the standard end-to-end model, and $P^{\text{gen}}_i$ is the generation probability. 
Details of the use of TCPGen in AED and RNN-T are given in the following sections.

\subsection{TCPGen in AED}

\begin{figure}[t]
    \centering
    \includegraphics[scale=0.36]{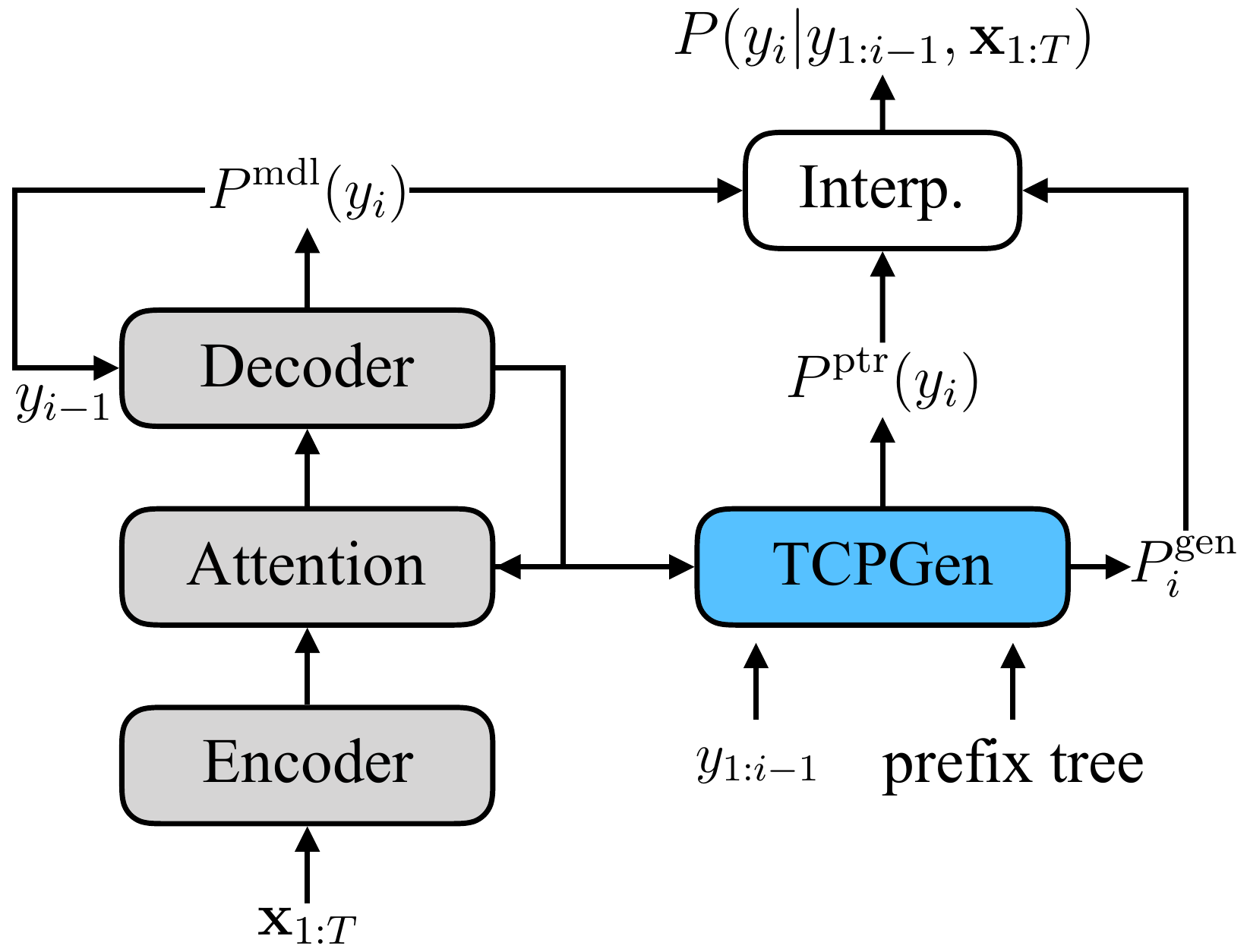}
    \vspace{-0.3cm}
    \caption{TCPGen component integrated in AED where Interp. corresponds to the interpolation in Eqn. \eqref{eq:TCPGen_final}.}
    \label{fig:TCPGen_in_las}
\end{figure}

TCPGen in AED is shown in Fig. \ref{fig:TCPGen_in_las}. Specifically, to calculate the TCPGen distribution, the query combines the context vector and the previously decoded token as shown in Eqn. \eqref{eq:TCPGeninlas_query}
\vspace{-0.1cm}
\begin{equation}
    \mathbf{q}_i = \mathbf{W}^{\text{Q}}_c\mathbf{c}_i + \mathbf{W}^{\text{Q}}_y\mathbf{y}_{i-1},
    \label{eq:TCPGeninlas_query}
    \vspace{-0.1cm}
\end{equation}
where $\mathbf{W}^{\text{Q}}_c$ and $\mathbf{W}^{\text{Q}}_y$ are parameter matrices. The history is limited to only the previous token to avoid the LM effect to bias towards common word piece patterns. Keys and values are computed from the decoder word piece embedding matrix 
as shown in Eqn. \eqref{eq:TCPGeninlas_key}.
\vspace{-0.1cm}
\begin{equation}
    \mathbf{k}_j = \mathbf{W}^{\text{K}}\mathbf{y}_j \hspace{1cm} \mathbf{v}_j = \mathbf{W}^{\text{V}}\mathbf{y}_j,
    \label{eq:TCPGeninlas_key}
\end{equation}
 where $\mathbf{y}_j$ denotes the $j$-th row of the embedding matrix. $\mathbf{W}^{\text{K}}$ and $\mathbf{W}^{\text{V}}$ are key and value parameter matrices which are shared throughout this paper. The TCPGen distribution and the TCPGen output can be computed using Eqns. \eqref{eq:TCPGen_attention} and \eqref{eq:TCPGen_value} respectively. The generation probability is calculated from the decoder hidden state $\mathbf{h}^{\text{dec}}_i$ and the TCPGen output $\mathbf{h}^{\text{ptr}}_i$, as shown in Eqn. \eqref{eq:las_gen}.
\begin{equation}
    {P}^{\text{gen}}_i = \sigma(\mathbf{W}^{\text{gen}}[\mathbf{h}^{\text{dec}}_i;\mathbf{h}^{\text{ptr}}_i]),
    \label{eq:las_gen}
\end{equation}
where $\mathbf{W}^\text{gen}$ is a parameter matrix. The distribution of $y_i$ can be calculated using Eqn. (\ref{eq:TCPGen_final}). In AED, DB can also be applied by combining a biasing vector with the decoder output before passing them to the Softmax output layer, as shown in Eqn. (\ref{eq:deepbiasing_las}).
\begin{equation}
    P^{\text{mdl}}(y_i) = \text{Softmax}(\mathbf{W}^{\text{O}} [\mathbf{h}^{\text{dec}}_i; \mathbf{c}_i] + \mathbf{W}^{\text{db}} \mathbf{h}^{\text{db}}_i),
    \label{eq:deepbiasing_las}
\end{equation}
where the biasing vector $\mathbf{h}^{\text{db}}_i$ is obtained from the sum of embeddings of all word pieces in $\mathcal{Y}_{\text{tree}}$, similar to \cite{deepshallow}. 

\subsection{TCPGen in RNN-T}
\begin{figure}[t]
    \centering
    \includegraphics[scale=0.36]{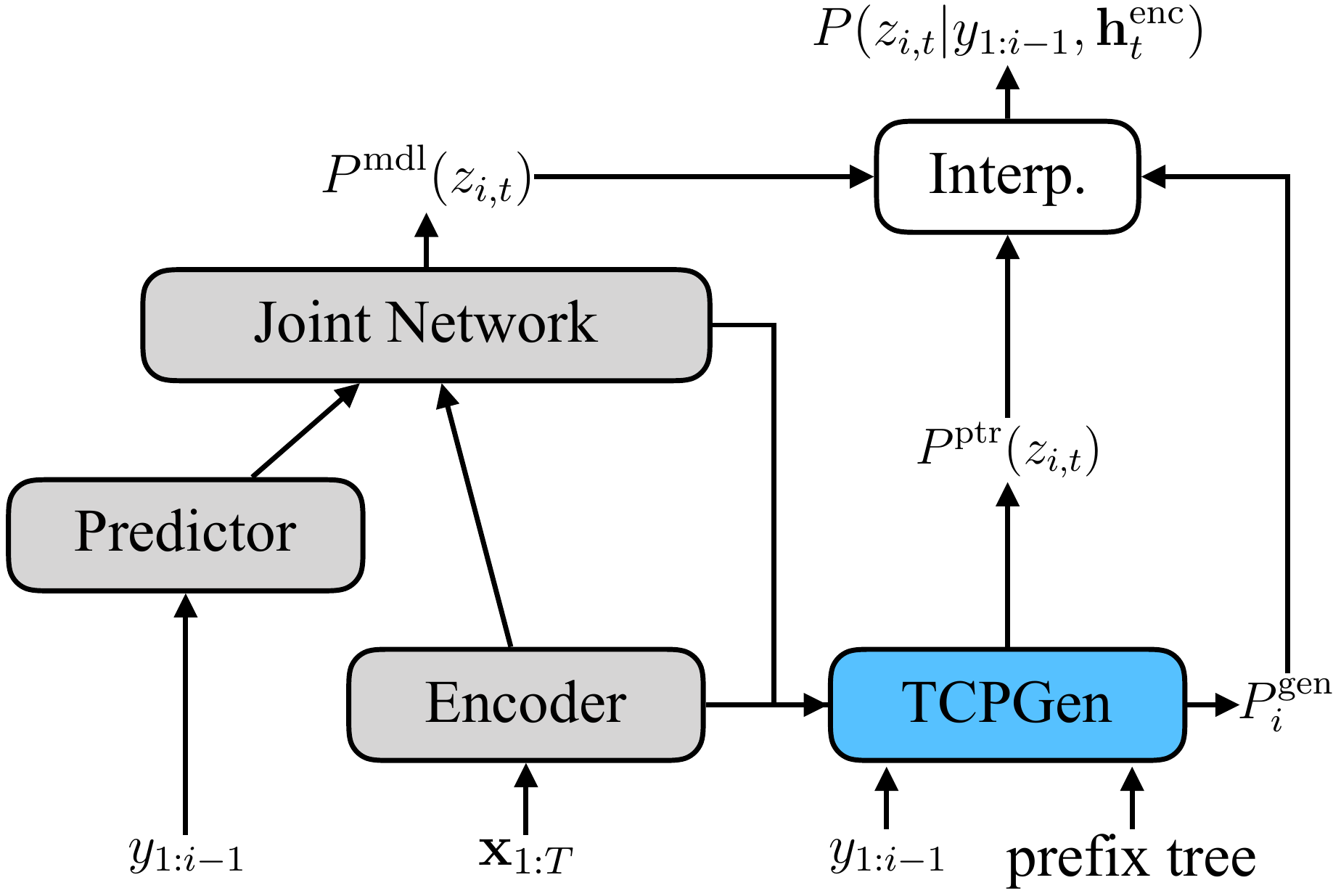}
    \vspace{-0.2cm}
    \caption{TCPGen component in the RNN-T model. The Interp. module corresponds to the interpolation in Eqns. \eqref{eq:TCPGen_final} and \eqref{eq:TCPGen_rnnt_interp}.}
    \vspace{-0.2cm}
    \label{fig:TCPGen_in_rnnt}
\end{figure}

TCPGen can be applied to RNN-T in a similar way as shown in Fig. \ref{fig:TCPGen_in_rnnt}, with differences in inputs and the distribution interpolation. 
Since there is no context vector in RNN-T, the query is derived from the encoder hidden state instead, as shown in Eqn. \eqref{eq:TCPGeninrnnt_query}. 
\begin{equation}
    \mathbf{q}_{i,t} = \mathbf{W}^{\text{Q}}_c\mathbf{h}^{\text{enc}}_t + \mathbf{W}^{\text{Q}}_y\mathbf{y}_{i-1}.
    \label{eq:TCPGeninrnnt_query}
\end{equation}
Note that the query vector is now specific to each encoder position $t$ and each predictor position $i$, and $\mathbf{y}_{i-1}$ is the word piece embedding from the predictor. Key and value vectors are derived from the predictor embedding matrix. The generation probability for each position $i, t$ is calculated using the penultimate layer output of the joint network and the TCPGen output vector. 
\begin{equation}
    {P}^{\text{gen}}_{i,t} = \sigma(\mathbf{W}^\text{gen}[\mathbf{h}^{\text{joint}}_{i,t};\mathbf{h}^{\text{ptr}}_{i,t}])
\end{equation}
As the $\varnothing$ symbol only exists in the model output distribution, the interpolation takes place only at word pieces as shown in Eqn. (\ref{eq:TCPGen_rnnt_interp})
\begin{equation}
P(z_{i,t}) = 
\begin{cases}
    P^{\text{mdl}}(\varnothing),& \text{if } z_{i,t} = \varnothing\\
    P(z_{i,t}),              & \text{otherwise}
\end{cases},
\label{eq:TCPGen_rnnt_interp}
\end{equation}
where $P(z_{i,t})$ is the interpolated probability in Eqn. (\ref{eq:TCPGen_final}) except that $P^{\text{ptr}}(z_{i,t})$ is scaled by a factor of $1-P^{\text{mdl}}(z_{i,t} \neq \varnothing)$ to ensure all probabilities sum to 1. 
Finally, DB can be applied by sending $\mathbf{h}^{\text{db}}_{i,t}$, or $\mathbf{h}^{\text{ptr}}_{i,t}$ if TCPGen is applied, to the input of the joint network, similar to the deep-shallow fusion in \cite{deepshallow}.

\section{Experimental Setup}
\label{sec:setup}
\subsection{Data}
The Librispeech corpus \cite{librispeech}, containing 960 hours of read English from audiobooks, was used to train and evaluate the proposed method. Models were trained on both the full 960-hour training data and the clean-100 split to show the effect on a relatively low-resource task. The dev-clean and dev-other sets were held out for validation, and test-clean and test-other used for evaluation. The 80-dim FBANK features at a 10~ms frame rate concatenated with 3-dim pitch features were used as inputs to models. SpecAugment \cite{specaug} with setting $(W,F,m_F,T,p,m_T)=(40,27,2,40,1.0,2)$ was used without any other data augmentation or speaker adaptation.

\subsection{Biasing list selection}
\label{sec:biasinglist}
Biasing lists were arranged at three different levels for evaluation, with corresponding specifications shown in Table \ref{tab:wlists}. The full rare word list containing 200k distinct words proposed in \cite{DBRNNT} was used as the collection of all biasing words. Following the scheme in \cite{DBRNNT}, biasing lists were first organised by finding words that belong to the full rare word list from the reference of each utterance and adding 1000 (or other amount specified in the experiments) distractors. This is referred to as the \textit{utterance-level biasing list} as the list is specific to a single utterance. 

Although the utterance-level biasing list was justified in \cite{DBRNNT} as a valid simulation for real tasks, it assumes that all biasing words in the test set will be covered during inference, and that there are no underlying relations between distractors. To alleviate these two assumptions, biasing lists together with distractors were obtained from two different ranges of text in the audiobook. In the first one, for each utterance, its chapter was located according to its chapter ID, and adjacent chapters combined until the length of the text reached 1000 lines which correspond to more than 10000 word tokens. The biasing list was made by extracting these 1000 lines and selecting distinct words in the biasing list which appeared in these text for each utterance, and the least frequent 1000 words kept to limit the size. Distractors were added for lists shorter than 1000 entries. The second one increased this extracting range to 10000 lines of text, followed by the same selection of biasing lists. When the boundary of a book is reached at one end, more lines were included from the other end to make sure the total amount of extracted text was similar, unless the book itself is shorter than 10000 lines. As the text length in the first arrangement is close to a book chapter, it is referred to as the \textit{chapter-level biasing list}. Meanwhile, the text length in the second one is close to a long novel and is referred to as the \textit{book-level biasing list}\footnote{{https://github.com/the-anonymous-bs/LibriSpeechBiasingLists.git}}. Furthermore, these two biasing list arrangements also simulate the situation where biasing words can be collected from a large amount of text-based knowledge, such as a lecture, a technical meeting or a conference.
\begin{table}[t]
    \centering
    \begin{tabular}{|l|c|}
    \hline
    Arrangement & Coverage (clean/other)\\
    \hline
    {Utterance-level biasing list}  & 10.7\% / 9.9\% \\
    {Chapter-level biasing list} &  10.6\% / 9.7\%\\
    {Book-level biasing list} & 2.9\% / 2.6\%\\
    \hline
    \end{tabular}
    \caption{Three different biasing list arrangements. Each list comprises 1000 words. Coverage is the total number of biasing word tokens divided by the total number of word tokens in each set.}
    \label{tab:wlists}
\end{table}

\subsection{System specifications and evaluation metrics}

Systems were built using the ESPnet toolkit \cite{espnet}. A unigram word piece model with 600 distinct word pieces was used. AED comprised a 4-layer Bi-LSTM-P encoder with 1024-dim hidden units projected to 2048-dim and a 2-block VGG frontend \cite{espnet, vgg}, a single-layer 1024-dim LSTM decoder and a 4-head 1024-dim location-sensitive attention. The RNN-T model had the same encoder, a 2-layer 1024-dim LSTM predictor and a 1024-dim joint network. TCPGen in both systems contained a 256-dim single-head attention layer and other layers were set by system dimensions. A 1024-dim embedding matrix was used for deep biasing. Moreover, a 2-layer 2048-dim LSTM-LM was trained on the Librispeech 800 million-word text training corpus and was used for shallow fusion. 

Systems trained with deep biasing or TCPGen used utterance-level biasing lists with 500 distractors for clean-100 experiments, and 1000 distractors for full 960-hour experiments. These lists were organised by finding biasing words from the reference and adding distractors. The dropping technique described in \cite{DBRNNT} was applied for training, with a drop rate of 40\% to prevent the model from being over-confident about TCPGen or deep biasing outputs. Beam search with a beam width of 30 was used for decoding, and a coverage penalty \cite{coverage_p} of 0.01 is applied for AED.

In addition to WER, a rare word error rate (R-WER) was used to evaluate the system performance on the biasing words which are ``rare" in the training data. R-WER is the total number of \textit{error} word tokens that belong to the biasing list divided by the total number of word tokens in the test set that belong to the biasing list. Insertion errors were counted into R-WER if the inserted word belonged to the biasing list\cite{DBRNNT}. 
Different biasing lists had different coverage which resulted in different denominators when calculating R-WER. Therefore, R-WER$_\text{u}$,R-WER$_\text{c}$, and R-WER$_\text{b}$ were used to denote R-WER for utterance, chapter and book-level biasing lists.
Moreover, as the number of biasing words in book-level lists is much smaller than the other two, a chapter-by-chapter sign test was performed to show the significance of R-WER improvements. The same significance test will also be applied to WER where the reduction is small.

\section{Experimental Results}
\label{sec:result}
\subsection{Librispeech clean-100 experiments}

\begin{figure}[t]
    \centering
    \includegraphics[scale=0.36]{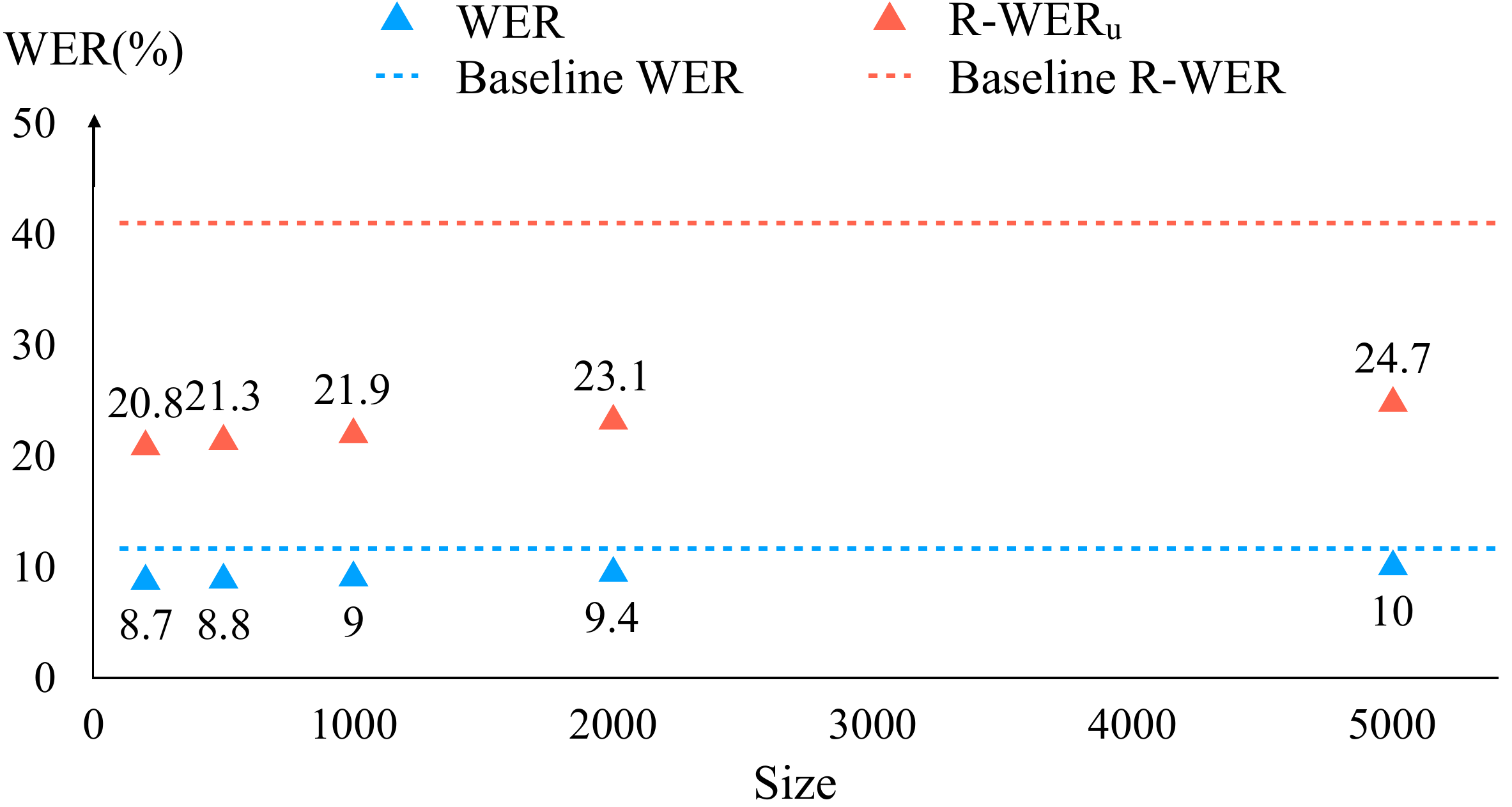}
    \vspace{-0.6cm}
    \caption{WER and R-WER$_\text{u}$ as a function of the number of distractors using utterance-level biasing lists. Baseline results correspond to the first row in Table \ref{tab:las_100}, and TCPGen results with 200, 500, 1000, 2000 and 5000 distractors are shown.}
    \label{fig:trend}
\end{figure}

In this section, systems are trained on the clean-100 training set and evaluated on the test-clean set. WERs and R-WERs evaluated with utterance-level and book-level biasing lists in AED are shown in Table \ref{tab:las_100}, where the baseline refers to the standard attention-based model without any biasing component. 
\begin{table}[t]
    \centering
    \begin{tabular}{|l|c|c|c|c|}
    \hline
    \multirow{2}{*}{System}  & \multicolumn{2}{c|}{Utterance-level} & \multicolumn{2}{c|}{Book-level} \\
    \cline{2-5}
       &  \%WER & \%R-WER$_\text{u}$ & \%WER & \%R-WER$_\text{b}$ \\
     \hline
     Baseline   & 11.6 & 40.9 & 11.6 & 59.6 \\
      + DB  & 9.6 & 27.2 & 11.4 & 48.9 \\
      + TCPGen  &   \textbf{9.0} & \textbf{21.6} & \textbf{10.9} & \textbf{34.1} \\
     \hline
    \end{tabular}
    \caption{WER and R-WER for attention-based model trained on clean-100 data and evaluated on the test-clean set. R-WER$_\text{u}$ and R-WER$_\text{b}$ denotes R-WER for utterance and book-level biasing lists respectively. DB used the sum of word piece embeddings.}
    \vspace{-0.3cm}
    \label{tab:las_100}
\end{table}
WER and R-WER reductions were achieved using deep biasing and TCPGen. In particular, TCPGen achieved a relative R-WER$_\text{u}$ reduction of 46.5\% compared to the baseline using utterance-level biasing lists.
When book-level lists were used, improvements in WER became smaller as the coverage of biasing lists was smaller according to Table \ref{tab:inc_coverage}. Besides, R-WER$_\text{b}$ was much higher than R-WER$_\text{u}$ since the biasing words covered by book-level lists are less frequent and tend to have a higher WER.
However, TCPGen still achieved a relative 6.0\% WER reduction. TCPGen with deep biasing did not provide extra improvements.
Furthermore, to examine the effect of increasing the number of distractors for AED, Fig. \ref{fig:trend} shows the the change in WER and R-WER$_\text{u}$ against the number of distractors using utterance-level biasing lists.
Up to 5000 distractors, TCPGen was still able to achieve improvements in WER and R-WER$_\text{u}$. Moreover, TCPGen also achieved high efficiency which is independent of the number of distractors, as the inference speed with 5000 distractors was the same as that with 1000 distractors, and was also close to that of the baseline.

\begin{table}[t]
    \centering
    \begin{tabular}{|l|c|c|c|c|}
    \hline
    \multirow{2}{*}{System}  & \multicolumn{2}{c|}{Utterance-level} & \multicolumn{2}{c|}{Book-level} \\
    \cline{2-5}
       &  \%WER & \%R-WER$_\text{u}$ & \%WER & \%R-WER$_\text{b}$\\
     \hline
     Baseline   & 13.2 & 42.7 & 13.2 & 58.9 \\
     +DB  & 12.9 & 38.6 & 13.2 & 54.5 \\
     +TCPGen  &  12.1 & 33.6 & 13.0 & 49.9 \\
     +TCPGen+DB  & \textbf{12.0} & \textbf{33.1} & \textbf{12.9} & \textbf{47.1} \\
     \hline
    \end{tabular}
    \caption{WER and R-WER for RNN-T trained on clean-100 data and evaluated on the test-clean set. DB uses $\mathbf{h}^{\text{db}}$ without TCPGen and $\mathbf{h}^{\text{TCPGen}}$ with TCPGen. R-WER$_\text{u}$ and R-WER$_\text{b}$ denotes R-WER for utterance and book-level biasing lists respectively.}
    \vspace{-0.3cm}
    \label{tab:rnnt_100}
\end{table}
Next, Table \ref{tab:rnnt_100} shows WER and R-WER using RNN-T with deep biasing and TCPGen components. In general, improvements using TCPGen in RNN-T were smaller than those in AED, which will be analysed in detail with experiments on the 960-hour training data set in the next section. The best performance was achieved by TCPGen combined with deep biasing, which resulted in a relative WER reduction of 8.3\% and a relative R-WER$_\text{u}$ reduction of 21.3\% with utterance-level lists, and a relative WER reduction of 2.3\% and a relative R-WER$_\text{b}$ reduction of 20.0\% with book-level lists.

\begin{table*}[t]
    \centering
    \begin{tabular}{|l|c|c|c|c|c|c|}
    \hline
    \multirow{2}{*}{System}  & \multicolumn{3}{c|}{Test-clean (\%)} & \multicolumn{3}{c|}{Test-other (\%)} \\
    \cline{2-7}
    & Utterance-level & Chapter-level & Book-level & Utterance-level & Chapter-level & Book-level \\
    \hline
     Baseline   & 4.4 (15.6) & 4.4 (15.1) & 4.4 (33.8) & 12.0 (35.8) & 12.0 (34.4) & \textbf{12.0} (60.3) \\
     Baseline + DB  & 4.0 (12.0) & 4.1 (12.1) & 4.4 (27.4$^*$) & 11.6 (30.2) & 11.6 (29.3) & 12.2 (51.6$^*$) \\
     Baseline + TCPGen  &  \textbf{3.7} (\textbf{8.3}) & \textbf{3.8} (\textbf{9.0}) & \textbf{4.3} (\textbf{20.6}$^*$) & \textbf{10.9} (\textbf{22.7}) & \textbf{11.3} (\textbf{23.9}) & 12.3 (\textbf{40.4}$^*$) \\
     \hline
     \hline
     Baseline + SF  & 3.7 (14.4) & 3.7 (13.8) & 3.7 (32.3) & 10.2 (32.9) & 10.2 (32.0) & {10.2} (58.3) \\
     Baseline + DB + SF & 3.4 (11.5) & 3.4 (11.6) & 3.6 (26.4$^*$) & 9.7 (28.1) & 9.8 (27.9) & 10.2 (52.0$^*$) \\
     Baseline + TCPGen + SF &  \textbf{3.0} (\textbf{8.3}) & \textbf{3.1} (\textbf{8.7}) & \textbf{3.5} (\textbf{19.4}$^*$) & \textbf{9.1} (\textbf{22.6}) & \textbf{9.4} (\textbf{23.2}) & \textbf{10.2} (\textbf{42.8}$^*$) \\
     \hline
    \end{tabular}
    \caption{WER and R-WER (in brackets) evaluated on test-clean and test-other sets for AED trained on 960-hour data. DB uses the sum of word piece embeddings. All three levels of biasing lists contain 1000 distinct word. SF stands for LM shallow fusion with a LM weight of 0.3. R-WER with $*$ indicates that the book-level R-WER reduction is significant ($p\leq 0.001$) compared to the baseline.}
    \label{tab:full_las}
\end{table*}

\begin{table*}[t]
    \centering
    \begin{tabular}{|l|c|c|c|c|c|c|}
    \hline
    \multirow{2}{*}{System}  & \multicolumn{3}{c|}{Test-clean (\%)} & \multicolumn{3}{c|}{Test-other (\%)} \\
    \cline{2-7}
    & Utterance-level & Chapter-level & Book-level & Utterance-level & Chapter-level & Book-level \\
    \hline
     Baseline   & 5.5 (18.7) & 5.5 (18.2) & 5.5 (37.9) & 15.3 (42.9) & 15.3 (41.2) & 15.3 (66.0) \\
     Baseline + DB  & 5.2 (15.2) & 5.3 (15.1) & 5.4 (31.5$^*$) & 14.6 (36.5) & 14.7 (34.9) & 15.2 (57.0$^*$) \\
     Baseline + TCPGen  &  {4.9} (13.9) & 5.1 (13.6) & 5.4 (28.2$^*$) & 14.0 (35.0) & 14.1 (32.4) & 14.8 (52.1$^*$) \\
     Baseline + TCPGen + DB  & \textbf{4.8} (\textbf{13.3}) & \textbf{4.8} (\textbf{12.8}) & \textbf{5.2} (\textbf{26.0}$^*$) & \textbf{13.9} (\textbf{33.5}) & \textbf{14.1} (\textbf{32.1}) & \textbf{14.8} (\textbf{51.2}$^*$) \\
     \hline
     \hline
     Baseline + SF  & 4.4 (15.6) & 4.4 (15.3) & 4.4 (33.1) & 12.5 (36.7) & 12.5 (35.7) & 12.5 (61.8) \\
     Baseline + DB + SF & 4.2 (13.1) & 4.3 (12.8) & 4.4 (28.7$^*$) & 12.1 (31.8) & 12.5 (31.9) & 12.5 (55.6$^*$) \\
     Baseline + TCPGen + SF & 4.0 (12.2)  & 4.1 (11.7) & 4.3 (26.0$^*$) & 11.6 (31.0) & 12.0 (\textbf{28.8}) & \textbf{12.1} (\textbf{49.7}$^*$) \\
     Baseline + TCPGen + DB + SF & \textbf{3.8} (\textbf{11.3}) & \textbf{4.0} (\textbf{11.0}) & \textbf{4.2} (\textbf{24.0}$^*$) & \textbf{11.5} (\textbf{29.0}) & \textbf{12.0} (29.3) & 12.2 (50.8$^*$) \\
     \hline
    \end{tabular}
    \caption{WER and R-WER (in brackets) evaluated on the test-clean and test-other sets for RNN-T trained on 960-hour data. DB uses $\mathbf{h}^{\text{db}}$ without TCPGen and $\mathbf{h}^{\text{ptr}}$ with TCPGen. All three levels of biasing lists contain 1000 distinct word. SF stands for LM shallow fusion with a LM weight of 0.3. R-WER with $*$ indicates that the book-level R-WER reduction is significant ($p\leq 0.001$) compared to the baseline.}
    \label{tab:full_rnnt}
\end{table*}

\subsection{Librispeech 960-hour experiments}

\subsubsection{Attention-based model trained on 960-hour data}

First, WER and R-WER using TCPGen in AED are shown in Table \ref{tab:full_las}. Results with SF are also provided. Using utterance-level biasing lists, AED with deep biasing and TCPGen achieved improvements in WER and R-WER$_\text{u}$ on both test-clean and test-other sets. The best performance in WER and R-WER$_\text{u}$ on both test-clean and test-other sets were obtained by the model with TCPGen, with 46.7\% relative R-WER$_\text{u}$ reduction on test-clean and 36.4\% relative R-WER$_\text{u}$ reduction on test-other. SF provided further WER reductions for all systems with sizeable improvements in common words and rather limited effects on biasing words. Therefore with SF, the model with TCPGen still achieved the best performance in both WER and R-WER$_\text{u}$, with a relative 42.2\% R-WER reduction on test-clean and a relative 34.9\% R-WER$_\text{u}$ reduction on test-other set. 

When chapter-level biasing lists were used, although WER and R-WER$_\text{c}$ for all biasing systems were lower than the baseline, both biasing methods showed performance degradation even though the coverage of biasing words are similar. This degradation came from morphologically similar words in the biasing list, such as \texttt{STRENGTH} to \texttt{STRENGTHEN}, that confused the biasing component. It was more likely to occur when the biasing list was extracted from text with contextual coherency where words were repeated in different forms. Besides, this phenomenon occurred more often in unbiased words for TCPGen when the reference word had another form in the biasing list, such as the example above where only \texttt{STRENGTHEN} was in the list. Despite this degradation, models with TCPGen still achieved the best performance among other models.

\begin{table}[t]
    \centering
    \begin{tabular}{|l|c|c|c|}
    \hline
     Size & 1000 & 2000 & 5000 \\
     Coverage & 2.6\% & 5.1\% & 9.4\% \\
     \hline
     Baseline \%WER & 10.2 (58.3) & 10.2 (42.9) & 10.2 (33.2) \\
     + TCPGen \%WER & \textbf{10.2} (\textbf{42.8}$^*$) & \textbf{10.0}$^*$ (\textbf{32.5}$^*$)& \textbf{10.2} (\textbf{26.1}$^*$) \\
     \hline
    \end{tabular}
    \caption{WER and R-WER$_\text{b}$ (in bracket) on test-other for AED with different book-level biasing list sizes and LM shallow fusion. WER with $*$ indicates a significant WER reduction at $p\leq 0.05$, and R-WER$_\text{b}$ with $*$ indicates a significant reduction at $p\leq 0.001$.}
    \label{tab:inc_coverage}
\end{table}
When book-level biasing lists were used, systems with biasing components experienced an increase in WER,
as the coverage of those biasing lists was much smaller. The reduction in error rates in biased words was offset by increased error rates in unbiased words, especially 
for those that were relatively rare in the training set. 
On test-clean, only TCPGen achieved a WER reduction, and the reduction became larger when shallow fusion was applied. On test-other, neither biasing method achieved a WER reduction without shallow fusion, as the coverage on test-other is even smaller. Shallow fusion mitigated the degradation in unbiased words, and TCPGen achieved WER reductions at a cost of slightly decreasing R-WER$_\text{b}$. 
Overall, TCPGen achieved the largest and statistically significant R-WER$_\text{b}$ reduction when book-level biasing lists were used, which also implied that the reduction is statistically significant for the other two levels where more biasing words were involved.

To increase the coverage of book-level biasing lists, the total number of distinct words in each biasing list was increased. Table~\ref{tab:inc_coverage} showed the change in coverage and WER when the size of biasing lists increased to 2000 and 5000. A consistent R-WER$_\text{b}$ reduction was observed as more biasing words were incorporated, 
since more biasing words were recognised correctly. As a result,  a 0.2\% absolute WER reduction was achieved when the size increased to 2000 and the coverage increased to 5.1\%. However, 
since many more distractors were added when the size increased to 5000,  performance degradation on unbiased words 
offset the improvement achieved on biasing words, resulting in a similar WER as 1000-word biasing lists.

\subsubsection{RNN-T trained on 960-hour data}
WERs and R-WERs for RNN-T with biasing components trained on the 960-hour data are shown in Table \ref{tab:full_rnnt}. As for AED, using biasing components achieved reductions in both WER and R-WER$_\text{u}$ when utterance-level biasing lists were used. When TCPGen was used in conjunction with DB, further WER and R-WER$_\text{u}$ reductions were achieved, with a relative 13.4\% WER reduction on test-clean, and a relative 9.1\% WER reduction on test-other compared to the baseline. Reductions on R-WER$_\text{u}$ were not as large as for AED, with a relative 28.8\% reduction on test-clean, and a relative 21.9\% reduction on test-other.


\begin{figure}[t]
    \centering
    \includegraphics[scale=0.37]{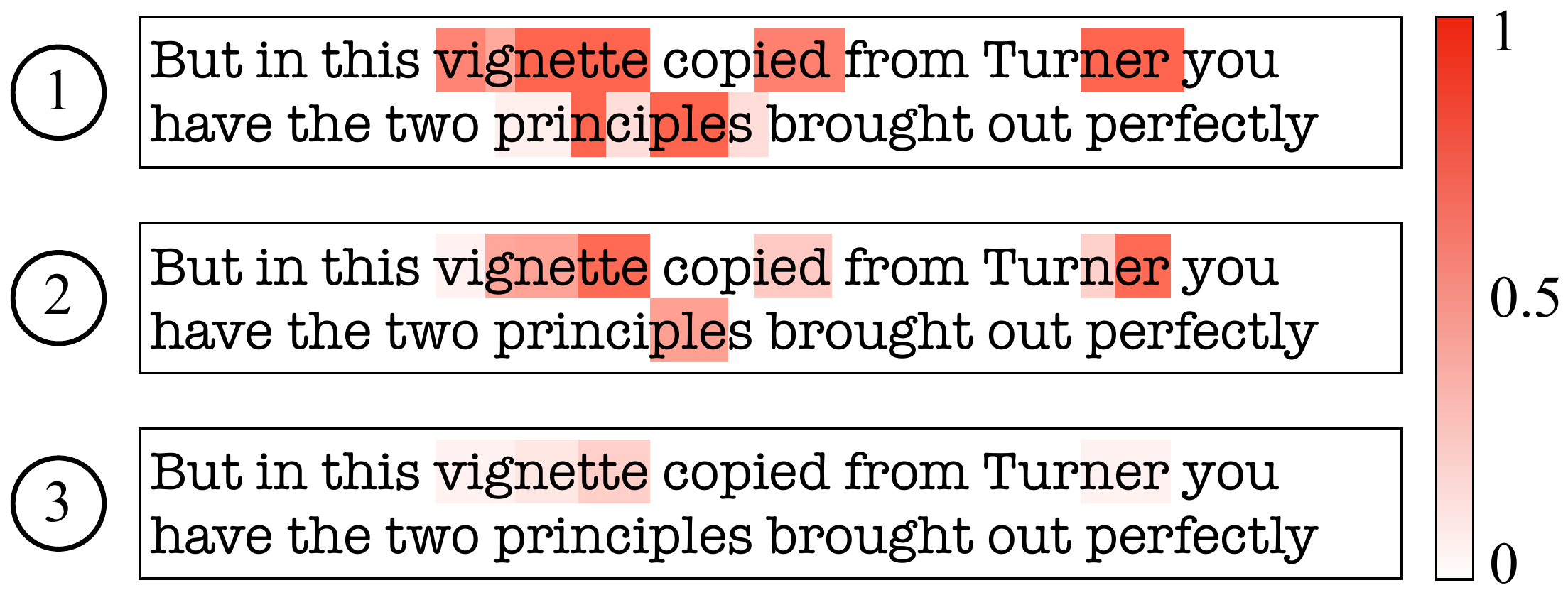}
    \caption{Heat map showing the generation probability for each word piece in an utterance taken from recognition results: \CircledText{1} AED + TCPGen; \CircledText{2} RNN-T + TCPGen; \CircledText{3} RNN-T + TCPGen + DB, to show how each system spots where to use contextual biasing. Biasing words are vignette and Turner.}
    \label{fig:example}
\end{figure}

As shown in Fig. \ref{fig:example}, RNN-T with TCPGen in general outputs a lower generation probability than AED with TCPGen. One possible reason to explain this difference is that the loss for RNN-T was calculated at each of the $T\times N$ combinations, but only a small portion of those correspond to outputting a new token that involved TCPGen. However, AED only calculated the loss at each decoder step, where TCPGen was always needed. As a result, RNN-T with TCPGen performed better on unbiased words while poorer on biased words than AED. Moreover, DB further reduced the dependency of RNN-T on the TCPGen component, as part of the functionality of the TCPGen component was undertaken by $\mathbf{h}^\text{ptr}$. A similar phenomenon was observed when using DB in AED, but the structural difference resulted in that improvements were only found in RNN-T.

When chapter-level and book-level biasing lists were used, a smaller degradation was observed in RNN-T since it relies less  on the TCPGen component. The best system on both test-clean and test-other sets using chapter and book-level biasing lists was the model with TCPGen, which also notably outperformed its DB counterpart. SF brought more improvements to RNN-T than to AED due to the different functionality of the predictor and the decoder \cite{stateless} in general. As before, similar relative improvements in R-WER with SF remained. Moreover, RNN-T with TCPGen performed better on biasing words than with TCPGen and deep biasing on the test-other set when SF was applied, as the effect of the TCPGen component was further diminished in the latter case.

\section{Conclusions}
\label{sec:conclusion}
This paper has proposed the tree-constrained pointer generator (TCPGen) component for contextual biasing in end-to-end ASR systems. TCPGen calculates the probability distribution over a subset of output subword units constrained by a prefix tree and interpolates this distribution and the model distribution using a dynamic factor. TCPGen was applied to AED and RNN-T. Experiments were performed on Librispeech audiobook data with biasing lists formulated at the utterance, chapter and book levels. Consistent and significant improvements in WER of biasing words were achieved by applying TCPGen in both end-to-end ASR models on the test-clean and test-other sets with biasing lists extracted at all three levels. 


\newpage
\begin{thebibliography}{8}

\bibitem{shallow_context_1}
I.~Williams, A.~Kannan, P.~Aleksic, D.~Rybach, T.~Sainath,
\newblock {``Contextual speech recognition in end-to-end
neural network systems using beam search"},
 \newblock  {\em Proc. Interspeech}, Hyderabad, 2018.

\bibitem{shallow_context_2}
Z.~Chen, M.~Jain, Y.~Wang, M.~L.~Seltzer \& C.~Fuegen
\newblock {``End-to-end contextual speech recognition using class language models and a token passing decoder"},
 \newblock  {\em Proc. ICASSP}, Brighton, 2019.
 
\bibitem{shallow_context_3}
D.~Zhao, T.~Sainath, D.~Rybach, P.~Rondon, D.~Bhatia, B.~Li \& R.~Pang,
\newblock {``Shallow-fusion end-to-end contextual biasing"},
 \newblock  {\em Proc. Interspeech}, Graz, 2019.
 
\bibitem{deep_context_1}
G.~Pundak, T.~Sainath, R.~Prabhavalkar, A.~Kannan \& D.~Zhao
\newblock {``Deep context: {E}nd-to-end contextual speech recognition"},
 \newblock  {\em Proc. ICASSP}, Calgary, 2018.
 
\bibitem{deep_context_2}
Z.~Chen, M.~Jain, Y.~Wang, M.L.~Seltzer \& C.~Fuegen
\newblock {``Joint grapheme and phoneme embeddings for contextual end-to-end ASR"},
 \newblock  {\em Proc. Interspeech}, Graz, 2019.

\bibitem{deep_context_3}
M.~Jain, G.~Keren, J.~Mahadeokar, G.~Zweig, F.~Metze \&
Y.~Saraf,
\newblock {``Contextual RNN-T for open domain ASR"},
 \newblock  {\em Proc. Interspeech}, Shanghai, 2020.
 
 \bibitem{deep_context_4}
U.~Alon, G.~Pundak \& T.~Sainath,
\newblock {``Contextual speech recognition with difficult negative training examples"},
 \newblock  {\em Proc. ICASSP}, Brighton, 2019.
 
 \bibitem{deep_context_5}
Z.~Chen, M.~Jain, Y.~Wang, M.~Seltzer \& C.~Fuegen
\newblock {``End-to-end contextual speech recognition using class language models and a token passing decoder"},
 \newblock  {\em Proc. ICASSP}, Brighton, 2019.

\bibitem{deepshallow}
D.~Le, G.~Keren, J.~Chan, J.~Mahadeokar, C.~Fuegen \& M.~L.~Seltzer
\newblock {``Deep shallow fusion for RNN-T personalization"},
 \newblock  {\em Proc. SLT}, 2021.
 
 \bibitem{DBRNNT}
D.~Le, M.~Jain, G.~Keren, S.~Kim, Y.~Shi, J.~Mahadeokar, J.~Chan, Y.~Shangguan, C.~Fuegen, O.~Kalinli, Y.~Saraf \& M.~L.~Seltzer
\newblock {``Contextualized streaming end-to-end speech recognition with trie-based deep biasing and shallow fusion"},
 \newblock  arXiv: {\em 2104.02194}, 2021.
 
\bibitem{word_mapping}
R.~Huang, O.~Abdel-hamid, X.~Li \& G.~Evermann
\newblock {``Class LM and word mapping for contextual biasing in end-to-end ASR"},
 \newblock  {\em Proc. Interspeech}, Shanghai, 2020.
 
\bibitem{dialogue_contextual}
Y.~Weng, S.~S.~Miryala, C.~Khatri, R.~Wang, H.~Zheng, P.~Molino, M.~Namazifar, A.~Papangelis, H.~ Williams, F.~Bell \& G.~Tur,
\newblock {``Joint contextual modeling for ASR correction and language understanding"},
 \newblock  {\em Proc. ICASSP}, Barcelona, 2020.
 
\bibitem{unsupervised_context}
Y.~M.~Kang \& Y.~Zhou
\newblock {``Fast and robust unsupervised contextual biasing for speech recognition"},
 \newblock  {\em arXiv:2005.01677}, 2020.
 
\bibitem{ne_correction}
A.~Garg, A.~Gupta, D.~Gowda, S.~Singh \& C.~Kim,
\newblock {``Hierarchical multi-stage word-to-grapheme named entity corrector for automatic speech recognition"},
 \newblock  {\em Proc. Interspeech}, Shanghai, 2020.
 
 \bibitem{lm_pointer}
D.~Liu, C.~Liu, F.~Zhang, G.~Synnaeve, Y.~Saraf \& G.~Zweig,
\newblock {``Contextualizing ASR lattice rescoring with hybrid pointer network language model"},
 \newblock  {\em Proc. Interspeech}, Shanghai, 2020.

\bibitem{audiovisual_context}
G.L.~Chao, C.~C.~Hu, B.~Liu, J.P.~Shen \& I.~Lane
\newblock {``Audio-visual TED corpus: enhancing the TED-LIUM corpus with facial information, contextual text and object recognition"},
 \newblock  {\em Proc. UbiComp}, 2021.

\bibitem{dialogue_contextual_2}
D.~Liu, C.~Liu, F.~Zhang, G.~Synnaeve, Y.~Saraf \& G.~Zweig,
\newblock {``Contextual language model adaptation for conversational agents"},
 \newblock  {\em Proc. Interspeech}, Hyderabad, 2018.
 
\bibitem{lecture_context}
M.B.~Andra \& T.~Usagawa,
\newblock {``Contextual keyword spotting in lecture video
with deep convolutional neural network"},
 \newblock  {\em Proc. ICACSIS}, Bali, 2017.


\bibitem{pointer_1}
A.~See, P.J.~Liu \& C.~D.~Manning
\newblock {``Get to the point: summarization with pointer-generator networks"},
 \newblock  {\em Proc. ACL}, Vancouver, 2017.
 
 \bibitem{pointer_2}
Z.~Liu, A.~Ng, S.~Lee, A.T.~Aw \& N.F.~Chen
\newblock {``Topic-aware pointer-generator networks for summarizing spoken conversations"},
 \newblock  {\em Proc. ASRU}, Singapore, 2019.
 
  \bibitem{pointer_3}
W.~Li, R.~Peng, Y.~Wang \& Z.~Yan
\newblock {``Knowledge graph based natural language generation with adapted pointer-generator networks"},
 \newblock  {\em Neurocomputing}, vol. 328, pp. 174--187, 2020.

\bibitem{e2e_attention_1}
J.~Chorowski, D.~Bahdanau, D.~Serdyuk, K.~Cho \& Y.~Bengio,
\newblock {``Attention-based models for speech recognition"},
 \newblock  {\em Proc. NIPS}, Montreal, 2015.
 
\bibitem{e2e_attention_2}
L.~Lu, X.~Zhang, K.~Cho \& S.~Renals,
\newblock {``A study of the recurrent neural network encoder-decoder for large vocabulary speech recognition"},
 \newblock  {\em Proc. Interspeech}, Dresden, 2015.
 
 \bibitem{e2e_attention_3}
D.~Bahdanau, J.~Chorowski, D.~Serdyuk, P.~Brakel \& Y.~Bengio,
\newblock {``End-to-end attention-based large vocabulary speech recognition"},
 \newblock  {\em Proc. ICASSP}, Shanghai, 2016.
 
  \bibitem{e2e_attention_4}
C.~C.~Chiu, T.~Sainath, Y.~Wu, R.~Prabhavalkar, P.~Nguyen, Z.~Chen, A.~Kannan, R.J.~Weiss, K.~Rao, E.~Gonina, N.~Jaitly, B.~Li, J.~Chorowski \& M.~Bacchiani,
\newblock {``State-of-the-art speech recognition with sequence-to-sequence models"},
 \newblock  {\em Proc. ICASSP}, Calgary, 2018.

 \bibitem{e2e_attention_5}
A.~Zeyer, K.~Irie, R.~Schlüter \& H.~Ney,
\newblock {``Improved training of end-to-end attention models for speech recognition"},
 \newblock  {\em Proc. Interspeech}, Hyderabad, 2018.
 
  \bibitem{e2e_attention_6}
Y.~Zhang, M.~Pezeshki, P.~Brakel, S.~Zhang, C.~Laurent Y.~Bengio, A.~Courville,
\newblock {``Towards end-to-end speech recognition with deep convolutional neural networks"},
 \newblock  {\em Proc. Interspeech}, San Francisco, 2016.
 
  \bibitem{e2e_attention_7}
T.~Hayashi, S.~Watanabe, Y.~Zhang, T.~Toda, T.~Hori, R.~Astudillo \& K.~Takeda,
\newblock {``Back-translation-style data augmentation for end-to-end ASR"},
 \newblock  {\em Proc. SLT}, Athens, 2018.
 
  \bibitem{e2e_attention_8}
S.~Karita, A.~Ogawa, M.~Delcroix, \& T.~Nakatani
\newblock {``Sequence training of encoder-decoder model using policy gradient for end-to-end speech recognition"},
 \newblock  {\em Proc. ICASSP}, Calgary, 2018.
 
 \bibitem{e2e_rnnt_1}
A.~Graves, A.~Mohamed \& G.~Hinton,
\newblock {``Speech recognition with deep recurrent neural networks"},
 \newblock  {\em Proc. ICASSP}, Vancouver, 2013.
 
\bibitem{e2e_rnnt_2}
R.~Prabhavalkar, K.~Rao, T.~Sainath, B.~Li, L.~Johnson \& N.~Jaitly,
\newblock {``A comparison of sequence-to-sequence models for speech recognition"},
 \newblock  {\em Proc. Interspeech}, Stockholm, 2017.
 
\bibitem{e2e_rnnt_3}
E.~Battenberg, J.~Chen, R.~Child, A.~Coates, Y.~Gaur, Y.~Li, H.~Liu, S.~Satheesh, D.~Seetapun, A.~Sriram \& Z.~Zhu,
\newblock {``Exploring neural transducers for end-to-end speech recognition"},
 \newblock  {\em Proc. ASRU}, Okinawa, 2017.
 
\bibitem{e2e_rnnt_4}
J.~Li, R.~Zhao, H.~Hu, Y.~Gong,
\newblock {``Improving RNN Transducer modeling for end-to-end speech recognition"},
 \newblock  {\em Proc. ASRU}, Singapore, 2019.
 
 \bibitem{e2e_rnnt_5}
Y.~He, T.~Sainath, R.~Prabhavalkar, I.~McGraw, R.~Alvarez, D.~Zhao, D.~Rybach, A.~Kannan, Y.~Wu, R.~Pang, Q.~Liang, D.~Bhatia, Y.~Shangguan, B.~Li, G.~Pundak, K.~C.~Sim, T.~Bagby, S.~Chang, R.~Rao \& A.~Gruenstein
\newblock {``Streaming end-to-end speech recognition for mobile devices"},
 \newblock  {\em Proc. ICASSP}, Brighton, 2019.

 \bibitem{stateless}
M.~Ghodsi, X.~Liu, J.~Apfel, R.~Cabrera \& E.~Weinstein
\newblock {``RNN-Transducer with stateless prediction network"},
 \newblock  {\em Proc. ICASSP}, Barcelona, 2020.

\bibitem{ISCA}
Q.~Li, C.~Zhang \& P.C.~Woodland,
\newblock {``Integrating source-channel and attention-based sequence-to-sequence models for speech recognition"},
 \newblock  {\em Proc. ASRU}, Singapore, 2019.
 
 \bibitem{e2e_ctc_1}
A.~Graves, S.~Fernandez, F.~Gomez, \& J.~Schmidhuber,
\newblock {``Connectionist temporal classification: labelling unsegmented sequence data with recurrent neural networks"},
 \newblock  {\em Proc. ICML}, Brighton, 2006.
 
  \bibitem{e2e_ctc_2}
A.~Zeyer, E.~Beck, R.~Schluter \& H.~Ney
\newblock {``CTC in the context of generalized full-sum HMM training"},
 \newblock  {\em Proc. Interspeech}, Stockholm, 2017.
 
  \bibitem{joint_ctc_atten}
S.~Kim, T.~Hori, \& S.~Watanabe,
\newblock {``Joint ctcattention based end-to-end speech recognition using multi-task learning"},
 \newblock  {\em Proc. ICASSP}, New Orleans, 2017.

\bibitem{SJTU}
Q.~Liu, Z.~Chen, H.~Li, M.~Huang, Y.~Lu \& K. Yu
\newblock {``Modular end-to-end automatic speech recognition framework for acoustic-to-word model"},
 \newblock  {in \em IEEE/ACM Transactions on Audio, Speech, and Language Processing}, vol. 28, pp. 2174-2183, 2020.
 
\bibitem{e2eLFMMI}
H.~Hadian, H.~Sameti, D.~Povey \& S.~Khudanpur,
\newblock {``End-to-end speech recognition using lattice-free MMI"},
 \newblock  {\em Proc. Interspeech}, Hyderabad, 2018. 

\bibitem{LAS}
W.~Chan, N.~Jaitly, Q.~V.~Le \& O.~Vinyals,
\newblock {``Listen, attend and spell: {A} neural network for large vocabulary conversational speech recognition"},
 \newblock  {\em Proc. ICASSP}, Shanghai, 2016.
 
 
 
 
 
 \bibitem{lmfusion_2}
A.~Kannan, Y.~Wu, P.~Nguyen, T.~Sainath, Z.~Chen \& R.~Prabhavalkar,
\newblock {``An analysis of incorporating an external language model into a sequence-to-sequence model"},
 \newblock  {\em Proc. ICASSP}, Calgary, 2018.
 
\bibitem{lmfusion_3}
T.~Hori, J.~Cho \& S.~Watanabe,
\newblock {``End-to-end speech recognition with word-based RNN language models"},
 \newblock  {\em Proc. SLT}, Athens, 2018.

 
\bibitem{lmfusion_5}
S.~Kim, Y.~Shangguan, J.~Mahadeokar, A.~Bruguier, C.~Fuegen, M.~L.~Seltzer \& D.~Le,
\newblock {``Improved neural language model fusion for streaming recurrent neural network transducer"},
 \newblock  {\em Proc. ICASSP}, Toronto, 2021.
 
 
 \bibitem{HAT}
E.~Variani, D.~Rybach, C.~Allauzen \& M.~Riley
\newblock {``Hybrid Autoregressive Transducer (HAT)"},
 \newblock  {\em Proc. ICASSP}, Barcelona, 2020.
 
\bibitem{ilme_1}
Z.~Meng, S.~Parthasarathy, E.~Sun, Y.~Gaur, N.~Kanda, L.~Lu, X.~Chen, R.~Zhao, J.~Li \& Y.~Gong,
\newblock {``Internal language model estimation for domain-adaptive end-to-end speech recognition"},
 \newblock  {\em Proc. SLT}, Shenzhen, 2021.
 
\bibitem{ilmt_1}
Z.~Meng, N.~Kanda, Y.~Gaur, S.~Parthasarathy, E.~Sun, L.~Lu, X.~Chen, J.~Li \& Y.~Gong,
\newblock {``Internal language model training for domain-adaptive end-to-end speech recognition"},
 \newblock  {\em Proc. ICASSP}, Toronto, 2021.
 
 
 \bibitem{specaug}
D.S.~Park, W.~Chan, Y.~Zhang, C.C.~Chiu, B.~Zoph, E.D.~Cubuk \& Q.V.~Le,
\newblock {``SpecAugment: {A} simple data augmentation method for automatic speech recognition"},
 \newblock  {\em Proc. Interspeech}, Graz, 2019.
 
  \bibitem{espnet}
S.~Watanabe, T.~Hori, S.~Karita, T.~Hayashi, J.~Nishitoba, Y.~Unno, E.~Y.~Soplin, J.~Heymann, M.~Wiesner, N.~Chen, A.~Renduchintala \& T.~Ochiai
\newblock {``ESPnet: {E}nd-to-end speech processing toolkit"},
 \newblock {\em Proc. Interspeech}, Hyderabad, 2018.
 
   \bibitem{vgg}
K.~Simonyan \& A.~Zisserman
\newblock {``Very deep convolutional networks for large-scale image recognition"},
 \newblock  {\em Proc. CVPR}, Columbus, 2014.
 
 \bibitem{librispeech}
V. Panayotov, G. Chen, D. Povey \& S. Khudanpur,
\newblock {``Librispeech: {A}n ASR corpus based on public domain audio books"},
 \newblock  {\em Proc. ICASSP}, South Brisbane, 2015.
 
 \bibitem{coverage_p}
J.~Chorowski \& N.~Jaitly
\newblock {``Towards better decoding and language model integration in sequence to sequence models"},
 \newblock  {\em Proc. Interspeech}, Stockholm, 2017.
 
\bibitem{transformer}
A.~Vaswani, N.~Shazeer, N.~Parmar, J.~Uszkoreit, L.~Jones, A.~N.~Gomez, L.~Kaiser \& I.~Polosukhin
\newblock {``Attention is all you need"},
 \newblock  {\em Proc. NIPS}, Long Beach, 2017.
 
\end{thebibliography}

\end{document}